# Deep-Ace: LSTM-based Prokaryotic Lysine Acetylation Site Predictor

Maham Ilyas[a], Abida Yasmeen[c], Yaser Daanial Khan[b], Arif Mahmood[a]

a) Department of Computer Science, Information Technology University (ITU), Lahore, Pakistan

b) Department of Computer Science, School of Systems and Technology, University of Management and Technology, Lahore, Punjab 54770, Pakistan.

c) Department of Chemistry, Lahore College for Women University, Lahore, Pakistan

**Abstract**

Acetylation of lysine residues (K-Ace) is a post-translation modification occurring in both prokaryotes and eukaryotes. It plays a crucial role in disease pathology and cell biology hence it is important to identify these K-Ace sites. In the past, many machine learning-based models using hand-crafted features and encodings have been used to find and analyze the characteristics of K-Ace sites however these methods ignore long term relationships within sequences and therefore observe performance degradation. In the current work we propose Deep-Ace, a deep learning-based framework using Long-Short-Term-Memory (LSTM) network which has the ability to understand and encode long-term relationships within a sequence. Such relations are vital for learning discriminative and effective sequence representations. In the work reported here, the use of LSTM to extract deep features as well as for prediction of K-Ace sites using fully connected layers for eight different species of prokaryotic models (including *B. subtilis, C. glutamicum, E. coli, G. kaustophilus, S. eriocheiris, B. velezensis, S. typhimurium,* and *M. tuberculosis*) has been explored. Our proposed method has outperformed existing state of the art models achieving accuracy as 0.80, 0.79, 0.71, 0.75, 0.80, 0.83, 0.756, and 0.82 respectively for eight bacterial species mentioned above. The method with minor modifications can be used for eukaryotic systems and can serve as a tool for the prognosis and diagnosis of various diseases in humans.

## 1. Introduction

The naturally occurring chemical protein modifications including methylation, acetylation, acylation, phosphorylation, glycosylation, ubiquitination, sulfation and farnesylation have been reported to influence the cell both in terms of its activity and stability. In addition to regulate different cellular mechanisms (gene expression regulation) these modifications also influence the other features of proteins including their proper folding for activity, interaction with other proteins and gene regulatory DNAs and RNAs and increasing their own stability [1,2].

Among almost 400 diverse modifications occurring in proteins the most of them occur on basic amino acids Lys, Arg and His. [3] Acetylation, acylation and SUMOylation are among 15 types of modifications occurring on lysine residues [1]. Acetylation of lysine residues (K-Ace) occurs post translationally in prokaryotes, usually the amino acids acetylated are internal in a peptide chain, while in eukaryotes both co-translational and post-translation events occur. Usually the post-translational modification (PTM) in prokaryotes results in 90% of their proteins acetylated at lysine residues internally. This internal PTM is reversible and may result in response to some drug metabolism [4]. The reaction is a covalent PTM catalyzed by lysine acetyltransferases (KATs), where "acetyl coenzyme A" being the acetyl group ($CH_3CO$) donor, while deacetylation is mediated by lysine deacetylases (KDATs) [5]. The cotranslational acetylation in eukaryotes is usually N-terminal acetylation, which is irreversible and is carried out by N-terminal acetylases. N-terminal acetylases however favor the acetylation of nonpolar aliphatic R groups containing amino acids (methionine, alanine) as well as polar uncharged R groups (serine, threonine) present on the N terminal. The presence of positively charged R groups of amino acids (Histidine, Lysine, Arginine) till at 5[th] position however interfere with this acetylation phenomenon at the termini [6].

Acetylation in prokaryotes is analogous to eukaryotic acetylation, with a concept that acetylation is highly conserved in evolution from prokaryotes to eukaryotes [7]. Since acetylation of internal lysines in a proteins is a reversible process and is a regulated process by KATs and KDATs (Fig. 1). The acetylation/ deacetylation regulation is important for the control of development and progression of certain diseases (cancer, congenital heart disorders (CHDs) and neurological disorders). Acetylation sometimes is crucial for the activity of some proteins e.g. P-53 protein (tumor suppressor protein) while deacetylation of tubulins and STAT3 proteins (signal transducer and activator of transcription 3) is essential for tumor control [8, 9, 10]. The K-Ace in certain Histone proteins have been reported for the proper development of heart, failure of which results in CHDs [11]. Similarly, deacetylation of sirtuin 1 protein protects the neurons from degeneration and eventually prevents neurodegeneration [12]. Multiple diseases and disorders have been linked to K-Ace dysregulation such as immune disorders, neurological diseases, cancer, and cardiovascular diseases [13-15]. A better understanding of the K-Ace in prokaryotes will ultimately lead to the understanding of the complex gene transcription events in humans and the mechanism cannot be denied for both prognosis and diagnosis of the so many disorders including those discussed above.

.

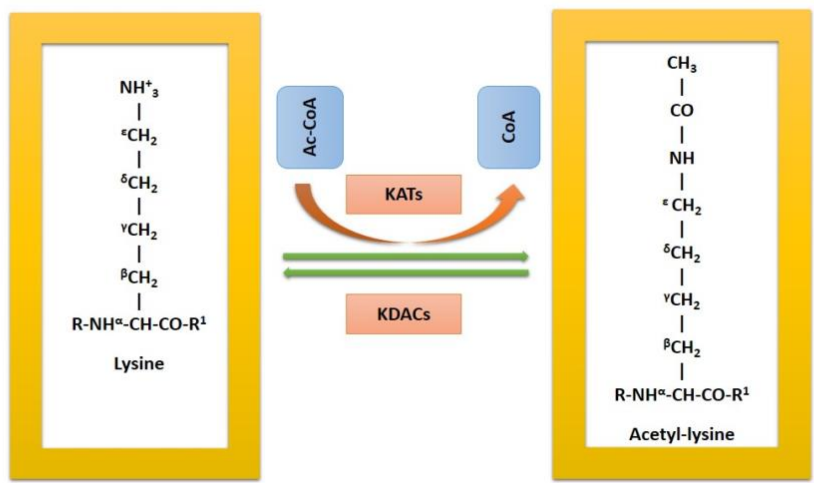

*Figure 1: Modification at Lysine residues supported by acetyltransferases, coenzyme A (CoA), acetyl coenzyme A (ac -CoA), lysine acetyltransferases (KATs), lysine deacetylases (KDACs)*

Due to significance of acetylation/ deacetylation in disease pathologies and cell biology, identification K-Ace sites and understanding its modulatory mechanism is quite important. In recent years many methods have been proposed such as mass spectrometry, chemical-based methods, and chromatin immune-precipitation to identify K-Ace PTM sites [16]. The latest technologies and tools for detection of K-Ace sites have been improved however when we consider the proteome size, we are able to observe only a slight segment of lysine 'modifyome'. As it's laborious to test every lysine residue in a single protein, hence this makes detection of K-Ace sites expensive and time-consuming with low throughput. Intensive work is required which creates a need for computational and efficient methods based on machine learning and deep learning. These methods have been proven to be efficient and provide accurate predictions. In the recent years, there have been multiple machine learning methods to identity K-Ace sites in prokaryotes and eukaryotes [17-21]. A typical flowchart of Prokaryotic Lysine Acetylation site prediction using deep learning is shown in Figure 2. Present tools for K-Ace prediction include prediction of Nε-acetylation on internal lysines (PAIL), prediction of lysine acetylation by support vector machines (LysAcet), prediction using an ensemble of support vector machine classifiers (EnsemblePail), solvent accessibility and physicochemical properties to identify protein N-acetylation sites (N-Ace), prediction of lysine methylation and lysine acetylation by combining multiple features (BPBPHKA), prediction based on multiple features (PLMLA), Proteome-wide analysis of amino acid variations that influence protein lysine acetylation (PSK-Ace Pred), lysine acetylation site prediction using logistic regression classifiers (K-Ace Pred), identifying acetylated lysine on histones and nonhistone proteins (LAceP), identification of species-specific acetylation sites by integrating protein sequence-derived and functional features (AceK), identifying multiple lysine PTM sites and their different types (SSPKA), species-specific lysine acetylation site prediction based on a large variety of features set (iPTM-mLys), prokaryote lysine acetylation sites prediction based on elastic net feature optimization (KA-predictor), functional analysis of prokaryote lysine acetylation site by incorporating six types of features into Chou's general (ProAcePred) and ProAcePred 2.0, prediction of human acetylation using a cascade classifier based on support vector machine and prediction of prokaryote lysine acetylation sites through deep neural networks with multi-information fusion (DNNAce) [22-38].

An ensemble method 'STALLION' which used eleven different encoding schemes has also been used previously. The concatenation of the extracted features is then used to finalize in three types of feature ranking methods. The designed five tree-based ensemble classifiers were later built species-specific. This method used handcrafted features which are time taking and do not guaranty optimal performance [37]. Even though there has been improvement in K-Ace sites prediction with different methods and techniques there still lies limitations among them as many state-of-the-art methods used hand crafted features and ML algorithms such as random forest (RF) and support vector machine (SVM) to train the model. With the development of Deep Learning (DL) models including CNN, LSTM and RNN, these can be utilized to compute feature subset which gives better representation of sequences rather than hand crafted features [38-43]. This may enhance the performance of K-Ace sites prediction and give more significant features than existing methods. Considering the limitations of the previous methods such as ProkAryotic, ProAcePred 2.0, and STALLION, we propose Deep-Ace which is state-of-the-art method for K-Ace site prediction in eight different prokaryotic species. Our proposed method is based on LSTM model for deep feature extraction due to its capability of encoding long term dependencies among sequences while previous methods were ML based. These features are more reliable than hand crafted features often used by the ML methods. LSTM helps us extract this information making our features more suitable for K-Ace site prediction.

## 2. Methodology

The methodology comprises four major steps which were (1) construction of benchmark data set, which comprises a total of 50,588 sequences of all species, (from which 11,685 are positive and 38,903 negative sequences); (2) extraction of deep features from sequences using deep neural networks; (3) classification using a well-known machine learning algorithm i.e., Random Forest (RF), AdaBoost (AB), gradient boosting (GB), extreme gradient boosting algorithm (XGB), extremely randomized tree (ERT) (4) validation of the proposed model using training, independent testing, 5-Cross validation, and 10-Cross validation (Figure 2).

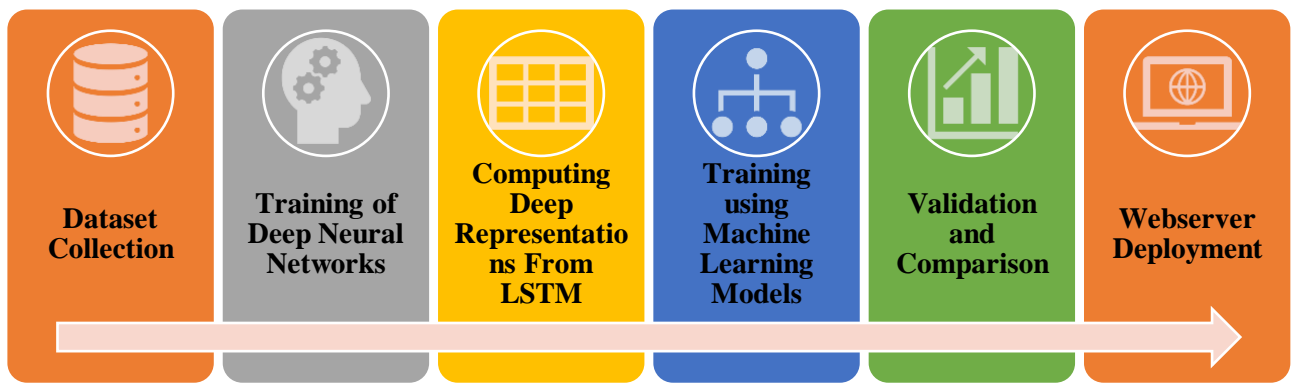

*Figure 2: A Typical Flowchart of Prokaryotic Lysine Acetylation Site Predictor using Deep Learning*

The extracted data from the PLMD database have been divided into eight species each having positive and negative samples. Then the samples were divided into the Training dataset and the Independent dataset. The sequences were used as raw input for the LSTM model to extract features. These features were then input into our ML models which classify our features into K-Ace and non-K-Ace sites (Fig. 3).

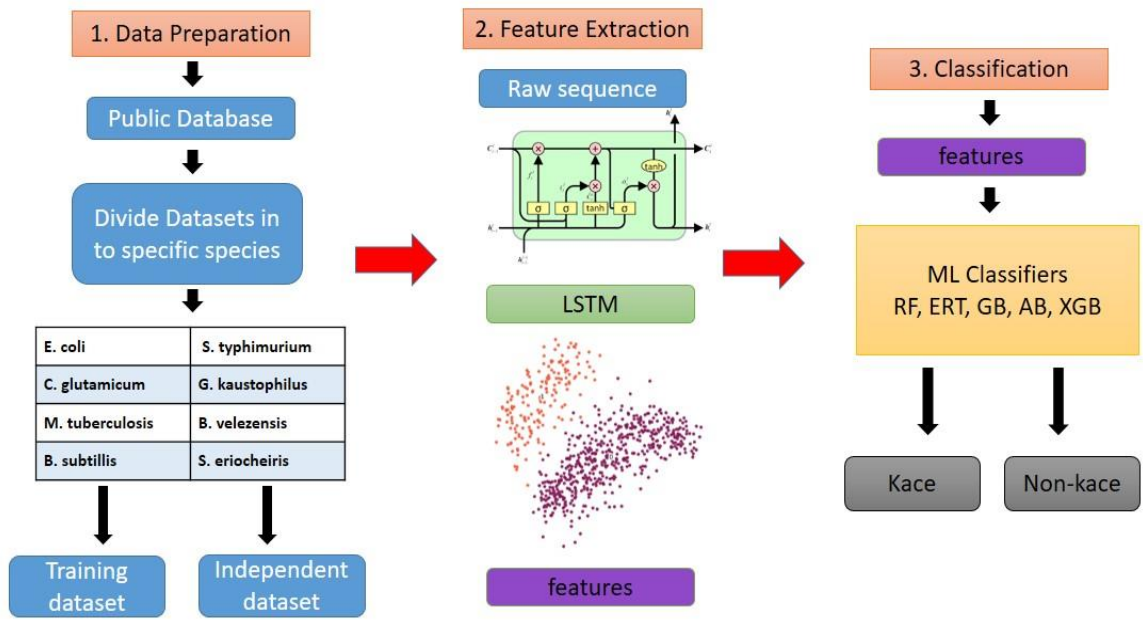

*Figure 3: An overview of the Deep learning-based predictor prokaryotic lysine acetylation sites. The three stages for prediction are shown.*

## 2.1. Data Collection

Based on the PLMD database (http://plmd.biocuckoo.org/download.php) novel redundant datasets were constructed by us after CD-Hit 30 for eight species *E. coli C. glutamicum, M. tuberculosis, B. subtilis,, S. typhimurium, G. kaustophilus, B. velezensis,* and *S. eriocheiris* [44, 45] . The optimal size with K at the center was 41 residue long sequences after experimenting with various fragment sizes. If K residue acetylation was found in the center those segments were labeled as positive samples (K-Ace) and otherwise labeled as negative samples (non-K-Ace). Hence dataset for this study was recently using rigorous methods to identify optimal length. A summary of K-Ace and non-K-Ace sequences have been shown in Table 1. Samples that are used for the training set and independent are imbalanced which gives us a better idea of how robust our model is, split we decided to do was 70% training data and 30% testing data.

*Table 1: A statistical summary of the training and independent datasets for eight species*

|  | Training | | Independent | |
|---|---|---|---|---|
| **Species** | **Positive** | **Negative** | **Positive** | **Negative** |
| *E. coli* | 3393 | 6698 | 1454 | 2872 |
| *C. glutamicum* | 701 | 2685 | 286 | 1166 |
| *M. tuberculosis* | 604 | 2607 | 259 | 1118 |
| *B. subtilis* | 903 | 3962 | 405 | 1681 |
| *S. typhimurium* | 109 | 883 | 52 | 374 |
| *G. kaustophilus* | 135 | 661 | 56 | 286 |
| *B. velezensis* | 1392 | 6653 | 579 | 2869 |
| *S. eriocheiris* | 951 | 3070 | 406 | 1318 |

### 3. Deep Neural Network (DNN)

In the past decades, deep neural networks (DNN) have been used to resolve various problems in different fields, similarly, they have helped solve bioinformatics problems as well [471]. The reason they became popular was their ability to extract useful information from data, which is a reliable and faster approach than extracting information using hand-crafted methods. These automatic methods of feature extraction combined with a machine learning classifier have proven to be effective [48]. The shallow layers extract low-level features and deeper layers extract high-level features hence optimizing even complex and non-linear functions give us state of art results [53]. These models use computational time and space effectively by doing a randomized search with hyperparameter tuning giving us optimal values of parameters that are optimal to our problem. DNNs have been effective in multiple fields but when it comes to being capable of processing sequences they cannot capture information in larger sequences and locate relationships between words hence to overcome this issue we have Recurrent Neural Network (RNN) [49, 50].

The reason they are called recurrent networks is that the same computation is performed on each element of the sequence and the output is dependent on the previous computation. They are effective due to their ability to operate over sequence vectors where the vector is given by $s_i = s_1, s_2, \ldots, s_n$, the computation is performed for a sequence of arbitrary length by using the recurrence formula $r_i = f_\emptyset(r_{i-1}, s_i)$, where $f_\emptyset$ represents Neural network and $\emptyset$ is a set of parameters, $i$ denotes steps taken to operate on every element of sequence hence having keeps the previous state combined with the current state keeping previous and future information of a sequence [51]. There are issues like vanishing gradient problems and shortage of memory where longer sequences could not be processed, hence the extension of RNN Long Short-Term Memory (LSTM) resolves these issues [52].

## 4. Long Short-Term Memory (LSTM) Network

To capture long-term dependencies in our sequences, we have employed the most popular cell structure Long Short-Term Memory (LSTM) for the current study. This architecture was proposed to mitigate the vanishing gradient problem by allowing the network to learn long-term dependencies [53, 54]. Whereas Figure 4 depicts the complete architecture of LSTM used in the current work to extract deep representations for the prediction of K-Ace sites. The ability of LSTM to resolve vanishing gradient problem enables it to handle noise and long-term dependencies to use larger sequences. The main functionality of the LSTM is given below:

$$\tilde{c}^{(t)} = \tanh(W^{(c)}x^{(t)} + U^{(c)}h^{(t-1)}) \qquad (1)$$

$$c^{(t)} = i^{(t)} \circ \tilde{c}^{(t)} + f^{(t)} \circ c^{(t-1)} \qquad (2)$$

$$h_t = o^{(t)} \circ \tanh(c^{(t)}) \qquad (3)$$

Where $c^{(t)}$ is the current memory contents computed over the previous activations time stamp $c^{(t-1)}$ and current memory content $\tilde{c}^{(t)}$. The $\tilde{c}^{(t)}$ is computed from the input sequence cell $x^{(t)}$ and the output of the previous hidden layer $h^{(t-1)}$. The current output of the hidden layer $h_t$ is computed from current candidate values $c^{(t)}$ and the output gate $o^{(t)}$ which is used to control the output of the hidden layer for the next state. The forget gate $f^{(t)}$ gets values from the previous state and current state and after activation, if the value is closer to 1 it retains information from previous candidate value $c^{(t-1)}$, and if closer to 0 forgets previous value and keeps the current value from $\tilde{c}^{(t)}$. The functionality of LSTM gates is given by the following equations:

$$o^{(t)} = \sigma(W^{(o)}x^{(t)} + U^{(o)}h^{(t-1)}) \qquad (4)$$

$$f^{(t)} = \sigma(W^{(f)}x^{(t)} + U^{(f)}h^{(t-1)}) \qquad (5)$$

$$i^{(t)} = \sigma(W^{(i)}x^{(t)} + U^{(i)}h^{(t-1)}) \qquad (6)$$

The input gate $i^{(t)}$ takes care if input information $x^{(t)}$ is relevant and can be added from the current state. LSTM units use a peephole which allows gated layers to look at the cell state [55]. These gates are controlled by learnable weights denoted by $W^{(o)}, W^{(f)}, W^{(i)}$ and $U^{(o)}, U^{(f)}, U^{(i)}$.

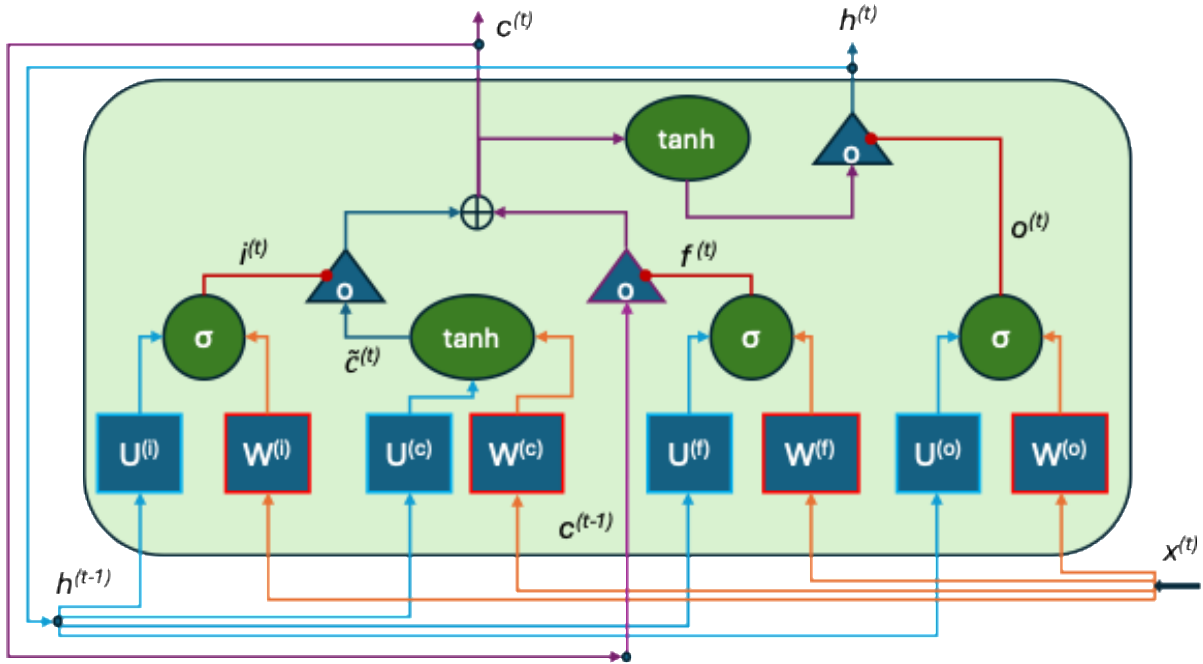

*Figure 4: LSTM Architecture*

## 4.1. Experimental Setup

According to our proposed methodology, for extracting deep features after multiple experiments we chose LSTM due to its performance and ability to understand dependencies in sequences. Our experimental setup is as follows early stopping was used in our training where if validation loss does not minimize after 3 consecutive epochs training will stop to avoid overfitting. As seen in Figure 5 the input layer has a dimension of (None, 41) where 41 is the length of a sequence given as input, and as batch size is 128 the input dimension becomes (128, 41). The next layer is the embedding layer where the dimension is (None, 41, 128) after this dropout layer is introduced which is kept at 0.2 which means 20% of neurons will randomly get disabled hence forcing all neurons to learn more effectively. The next layer is LSTM where the dimension changes to (None, 64) after adding another dropout we have our last layer where the dimension is as (None, 1) with the optimizer as adam and the activation function as sigmoid. We have extracted features from this network to take as input for ML classification models where they predict K-Ace and non-K-Ace sites. These features are called LSTM representation and are extracted from the dropout_1 layer hence our input size for ML classifiers is (1, 64).

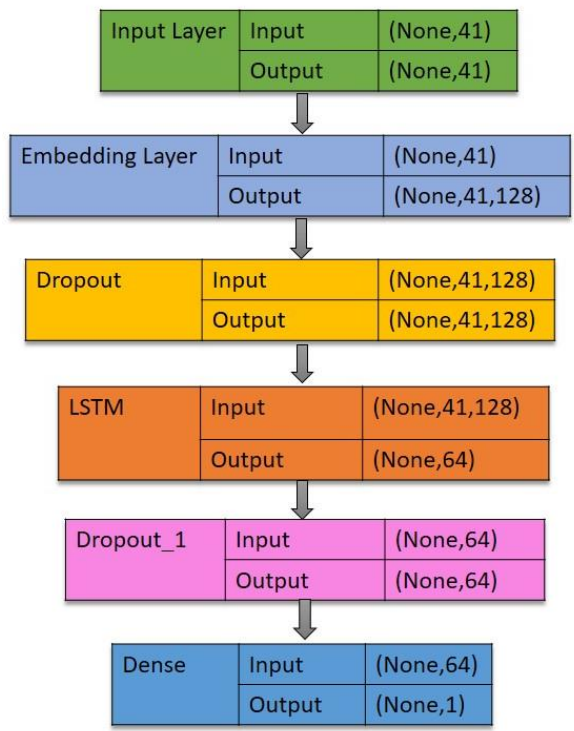

*Figure 5: Summary of the proposed LSTM Model*

## 5. Machine Learning Algorithms for Classification

We have used multiple classifiers to have a comprehensive comparison of the features extracted by using LSTM architecture. These classifiers help us to evaluate the performance, each of them is explained in the following section:

### 5.1.  Random Forest

Random Forest (RF) is a supervised ML algorithm, and it's been widely used for classification and regression problems [56]. We used it for binary classification which means when our output is 1 it means the sequence is K-Ace sites and when it is 0 it means non-K-Ace sites [57]. RF has been used previously in acetylation protein identification and recognizing protein-metal ion ligands binding residues [58] [59]. The major benefits offered by this model are the diversity of each tree, reduced dimensionality due to non-consideration of all the features, parallelization, fix train-test split ratio, bagging, and stability. To verify the accuracy of the RF model, a method called out-of-bag estimate (OOB) is employed by RF to measure the prediction error of the algorithm. These OOB samples help in achieving an unbiased balance of separation error. The structure of the random forest algorithm is presented in Figure 6.

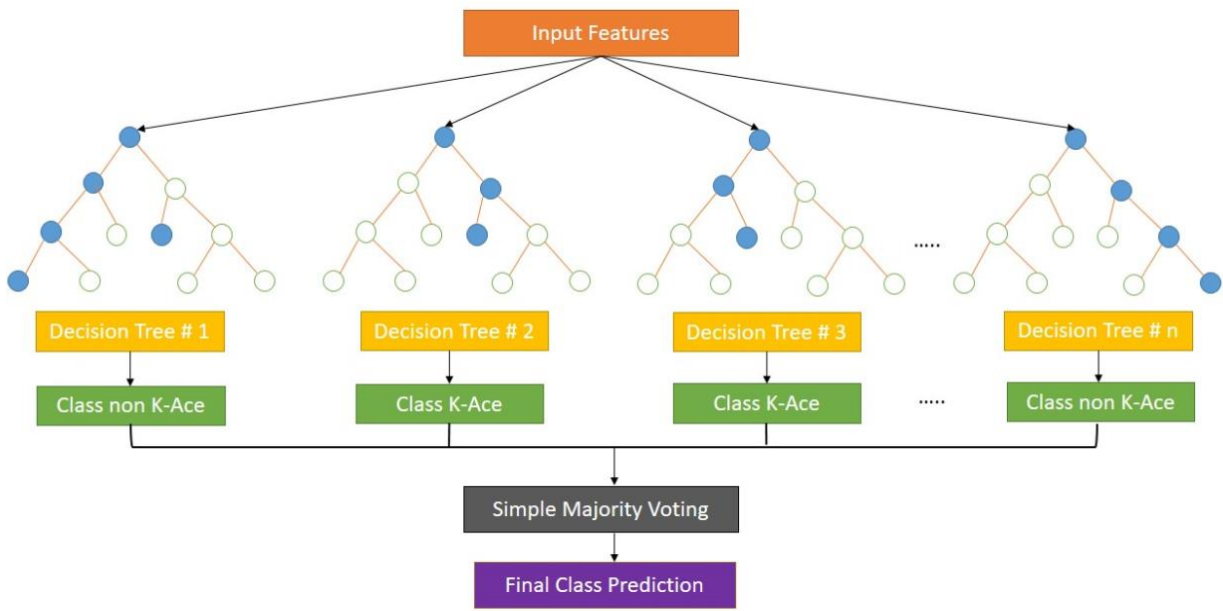

*Figure 6: The Structure of the Random Forest (RF) Model*

## 5.2. Extremely Random Tree

An Extremely randomized tree (ERT) method is used for both regression and classification problems. It has been widely used in protein-to-protein interaction predictions [60]in Identifying promoters and their strength. It is capable of efficiently randomizing attributes and during splitting tree nodes it makes more accurate cut-points. The main contribution of ERT is that the trees are built randomly and are independent of output values. Randomization can be tuned by changing parameters accordingly to the problem we are dealing with. Other than accuracy ERT is computationally efficient [61].

## 5.3. AdaBoost

Previously AdaBoost (AB) has been used in many protein subcellular localization classification and acetylation predictions [62, 63]. AB is used in ML as an ensemble method where weights are re-assigned to each instance, initially, all weights are assigned equally but for every instance that is predicted wrong, a higher weight is assigned to it. Hence in another cycle algorithm tries to learn wrongly predicted instances. It reduces bias and variance and sequentially improves results.

## 5.4. Gradient Boosting

Friedman introduced Gradient boosting (GB) made from boosting and gradient descent having a combination of three components loss function, additive model, and a weak learner [64]. GB has been efficient in DNA binding classification and prediction of hot spots at protein-protein interfaces [65, 66]. It differs from AB by utilizing a differentiable loss function making it more stable upon meeting outliers. It also predicts error which was left by previous models utilizing a loss optimization function through the gradient descent method.

## 5.5. Extreme Gradient Boosting

Extreme Gradient Boosting (XGB) is more efficient and scalable than GB for building trees. The method used to determine the best node splits are gain and similarity score. XGB has been widely used such as in predicting the origin of replication sites in DNA [46], predicting protein submitochondrial localization [67], and computational prediction of druggable proteins using extreme gradient boosting [68].

## 6. Results and Discussion

### 6.1. Evaluation Metrics

To measure the performance of computational models researchers have developed various quantitative metrics which help us compare how well models predict using different aspects of validations [69]. To compare different models and decide which gives us better results for K-Ace site prediction and analyze experimental results for validation and verification we use one of the most widely used metrics Classification accuracy (CA). Where when CA is equal to 1 means no sequence in our data was predicted incorrectly i.e. $FN = TN = 0$ and where $CA = 0$ which means we have no K-Ace sites predicted correctly. The formula can be seen in equation (7) it tells us how many correct predictions we have made over the total prediction made [70].

$$CA = \frac{(TP + TN)}{P + N} \qquad (7)$$

Where P is positive, and N is negative samples and P+N represents the number of test samples. True positives (TP) represent the correctly predicted number of K-Ace sites and TN correctly predicted non-K-Ace sites predicted. Precision known as Specificity (Sp) focuses on negative samples and how many sites are non-K-Ace sites in our study similarly Recall known as Sensitivity (Sn) focuses on positive samples that how many sites are K-Ace sites. False positive (FP) represents the number of sequences that were non-K-Ace and predicted as K-Ace. False-negative (FN) represents the number of non-K-Ace sequences which were predicted as K-Ace sequences similarly.

$$Sn = Recall = \frac{TP}{(TP + FN)} \qquad (8)$$

$$Precision = \frac{TP}{(TP + FP)} \tag{9}$$

$$Sp = \frac{TN}{(FP + TN)} \tag{10}$$

Let's understand what their values mean during an evaluation when we have zero samples that were wrongly predicted as non-K-Ace, then our Sn will be maximum hence $TN = 0$, similarly, when we have $FN = 0$ our Sp will be maximum as seen in equation (8) and (10) mentioned above. The Area under the curve (AUC) measures the ability of the prediction model which helps us distinguish between negative and positive class. High values mean our model can differentiate between the K-Ace and non-K-Ace sites hence giving us a summary of the receiver operating characteristic (ROC) curve [71]. ROC is a graphical representation of the false positive rate and true positive rate of our model hence telling us a performance on all thresholds of classification.

$$F1\ Score = \frac{2TP}{(2TP + FP + FN)} \tag{11}$$

F1 score or F-measure [72] gives us a weighted average and balanced overview of our precision and recall as to when we have to look at our precision and recall both together taking a simple average one value can undermine others giving us a false estimation. It uses precision value which quantifies positive class predictions telling us the closeness of two or more measurements. It gives us an estimation of how our model performs in both positive and negative classes as seen in the above equation (11).

$$MCC = \frac{(TP \times TN) - (FP \times FN)}{\sqrt{(TP + FP) \times (TP + FN) \times (TN + FP) \times (TN + FN)}} \tag{12}$$

Mathew's correlation coefficient factor (MCC) is used when we have an imbalanced dataset such as in our case where negative samples are more than positive samples [73]. A model is most accurate when $MCC = 1$ and when $MCC = -1$ it means model has performed poorly. As we are working with binary classification hence, we already have a 50% probability of predicting correctly before training due to the availability of two classes positive and negative. So any random guess made by predicting model will result in a positive or negative class giving us 50% correct results. In our study, if $MCC = 0$ and $CA = 0.5$ where $TN = \frac{TP}{2}$ and $FN = \frac{FP}{2}$ tells us that only half of the non-K-Ace and half of the K-Ace sequences were predicted accurately. Hence these accuracy measures help us analyze the capability of the model and understand the relationship between our samples. Hence for fair evaluation, we will be reporting all these metrics on training, independent testing, 5-Cross validation, and 10-Cross validation.

## 6.2. Accuracy Estimation Through Training

We evaluate our proposed algorithm and existing methods by using different performable measures discussed in 6.1. The available dataset is randomly split into 70-30% where 70% is used for training and 30% for testing. For all ML classifiers, we used LSTM representations as input discussed in section 4.1, where LSTM is ending output of the LSTM model. For overall comparison of all different species, the average performance is reported in Table 2.

*Table 2: Performance comparison of the proposed feature on the training set using different classifiers including Long short-term memory (LSTM) and ML classifiers Random Forest (RF), Extreme Random Tree (ERT), Gradient Boosting (GB), AdaBoost (AB), Extreme Gradient Boosting (XGB)*

| Species | Classifier | CA | Sn | Sp | MCC | AUC | F1 |
|---|---|---|---|---|---|---|---|
| *Average of all species* | LSTM | 0.97 | 0.99 | 0.88 | 0.89 | 0.94 | 0.98 |
| | RF | 1 | 1 | 1 | 1 | 1 | 1 |
| | ERT | 1 | 1 | 1 | 1 | 1 | 1 |
| | GB | 0.97 | 0.99 | 0.9 | 0.9 | 0.98 | 0.99 |
| | AB | 0.98 | 0.99 | 0.94 | 0.93 | 0.99 | 0.99 |
| | XGB | 0.99 | 1 | 0.94 | 0.94 | 1 | 0.99 |

As can be observed in Table 2 features extracted through LSTM and classified using different Machine Learning classifiers give us outstanding results. The training graphs for every species with their respective ROC curve for each classifier can be seen in Figure 7.

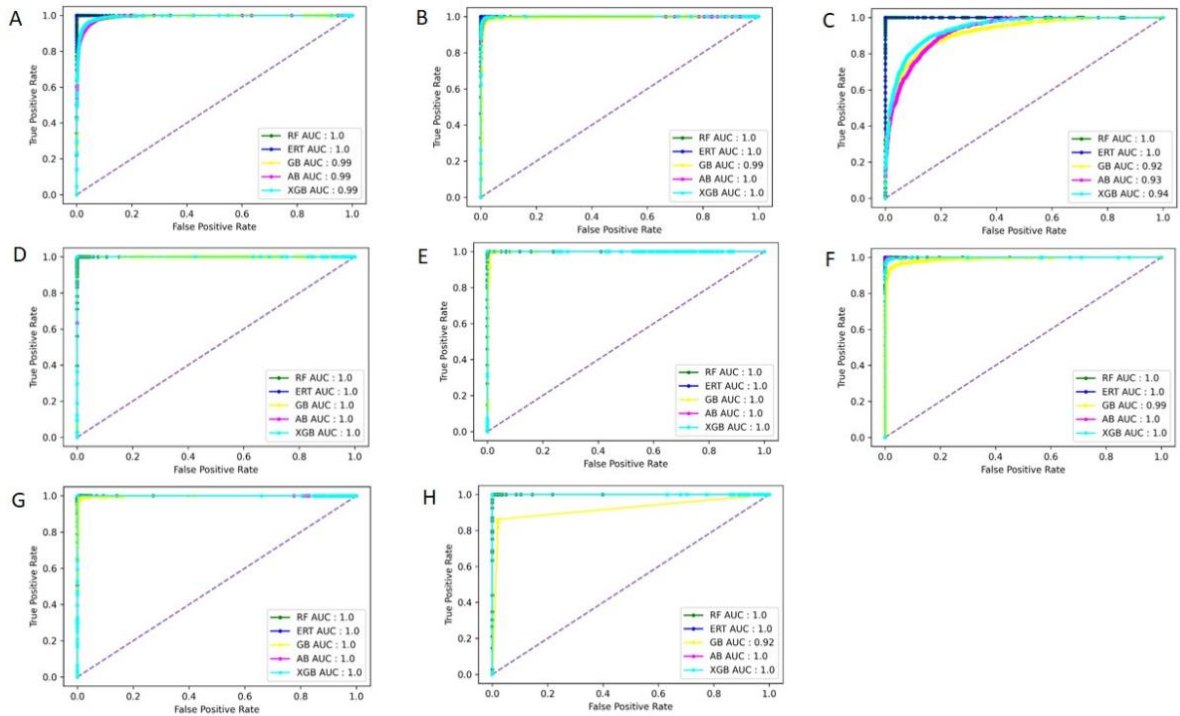

*Figure 7 ROC curve of ML Classifiers on training set, A: E. coli, B: B. velezensis, C: Bacilus subtilis, D: C. glutamicum, E: G. kaustophilus, F: M. tuber, G: S. eriocheiris, H: S. typhimurium*

## 6.3. Accuracy Estimation Through Independent Testing

Once the model has been trained comes the most trivial test considered for prediction sequence models known as independent dataset testing where positive and negative samples which our model has not seen are predicted to analyze the true potential of the model and understand its capability to distinguish between classes. 30% of samples were taken as independent comprising 15181 sequence segments. Results on metrics mentioned in 6.1 earlier are given below in Table 3. As you can see our method has outperformed for every species and the ML classifiers trained on LSTM representation have given us state-of-the-art results for all species.

*Table 3: Performance comparison of the proposed feature on independent set using different classifiers including Long short-term memory (LSTM) and ML classifiers Random Forest (RF), Extreme Random Tree (ERT), Gradient Boosting (GB), AdaBoost (AB), Extreme Gradient Boosting (XGB)*

| Species | Classifier | CA | Sn | Sp | MCC | AUC | F1 |
|---|---|---|---|---|---|---|---|
| *Average of all species* | DNN | 0.79 | 0.88 | 0.35 | 0.25 | 0.62 | 0.87 |
| | RF | 0.79 | 0.89 | 0.36 | 0.26 | 0.71 | 0.87 |
| | ERT | 0.79 | 0.89 | 0.35 | 0.25 | 0.71 | 0.87 |
| | GB | 0.77 | 0.87 | 0.38 | 0.24 | 0.69 | 0.86 |
| | AB | 0.78 | 0.87 | 0.38 | 0.25 | 0.71 | 0.86 |
| | XGB | 0.78 | 0.87 | 0.37 | 0.25 | 0.72 | 0.86 |

The Independent graphs for every species with their respective ROC curve for each classifiers can be seen in the Figure 8 below.

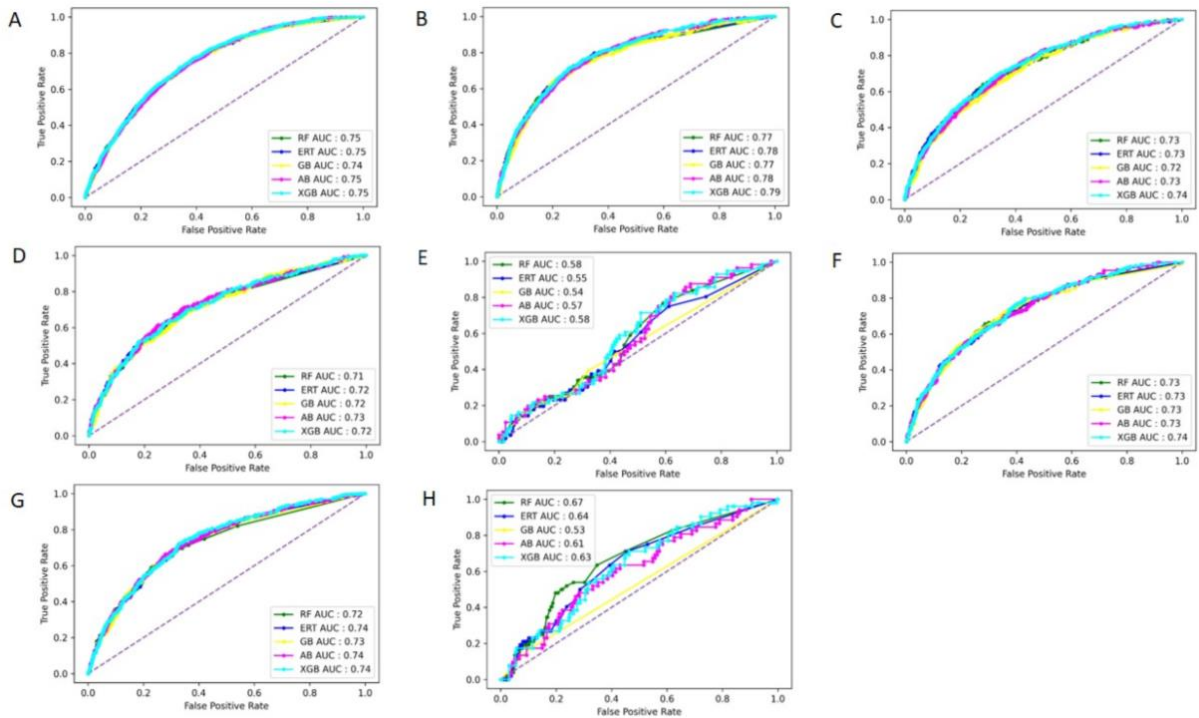

*Figure 8 ROC curve of ML Classifiers on independent set, A: E. coli, B: B. velezensis, C: Bacilus subtilis, D: C. glutamicum, E: G. kaustophilus, F: M. tuber, G: S. eriocheiris, H: S. typhimurium*

## 6.4. Accuracy Estimation Through Cross Fold Validation

A technique for robust analysis of a model is a cross-validation test or known as k-fold cross-validation test consisting of sub-sampling [74]. As sampling of train and test samples in each fold is unknown to the model and has diverse and biased in the case of random disjoint k partitioning we analyze the robustness of the model for predicting sequences [75, 76]. It applies independent data set testing which overcomes shortcomings of the training test i.e. we have unknown data for finding comparative performance of the different models. [39, 76, 77]. As each sequence has the opportunity to be part of training and testing data which makes models robust and handles out-of-sample data more efficiently giving us an unbiased picture of our predictor. In k-fold cross-validation, k represents the number of times different partitions of subsamples will be done as training data (i.e., k-1 partitions) shown in Figure 24.

*Figure 9 Structure of 10-Fold Cross-Validation*

Further, to find the model's accuracy average of each k-test result is computed. 10-fold cross-validation technique representation is shown in Figure 9 where k = 10 divides the whole data set into 10 disjoint sets (i.e., splits or folds). In this study, we have computed an average for all the splits using 10-fold cross-validation where k=10 which means 90% is training and 10% is test set. Experiments are also repeated using 5-fold cross-validation where k = 5 which means 80% is training and 30% is test set. Data samples will be randomly selected to make the number of folds given where each split is predicted by the model and an average is calculated for all the splits as 10-fold cross-validation is shown in Table 4. Figure 10 is the graphical representations of 10-fold cross-validation ROC curve computed for LSTM using 10-fold cross-validation. 5-fold cross-validation shown in Table 5, followed by Figure 11 is the graphical representations of the 5-fold cross-validation ROC curve computed for LSTM respectively.

*Table 4: Performance comparison of the proposed feature on 10-fold cross-validation set using different classifiers including Long short term memory (LSTM) and ML classifiers Random Forest (RF), Extreme Random Tree (ERT), Gradient Boosting (GB), AdaBoost (AB), Extreme Gradient Boosting (XGB)*

| Species | Classifier | ACC | SN | SP | MCC | AUC | F1 |
|---|---|---|---|---|---|---|---|
| *Average of all species* | LSTM | 0.79 | 0.88 | 0.35 | 0.25 | 0.62 | 0.87 |
| | RF | 1 | 1 | 1 | 1 | 1 | 1 |
| | ERT | 0.97 | 0.99 | 0.87 | 0.88 | 1 | 0.98 |
| | GB | 0.95 | 0.98 | 0.8 | 0.81 | 1 | 0.97 |
| | AB | 0.97 | 0.99 | 0.9 | 0.9 | 1 | 0.98 |
| | XGB | 0.96 | 0.99 | 0.84 | 0.85 | 1 | 0.98 |

10-fold cross-validation graphs for every species with their respective ROC curve for each classifiers can be seen in the Figure 10 below.

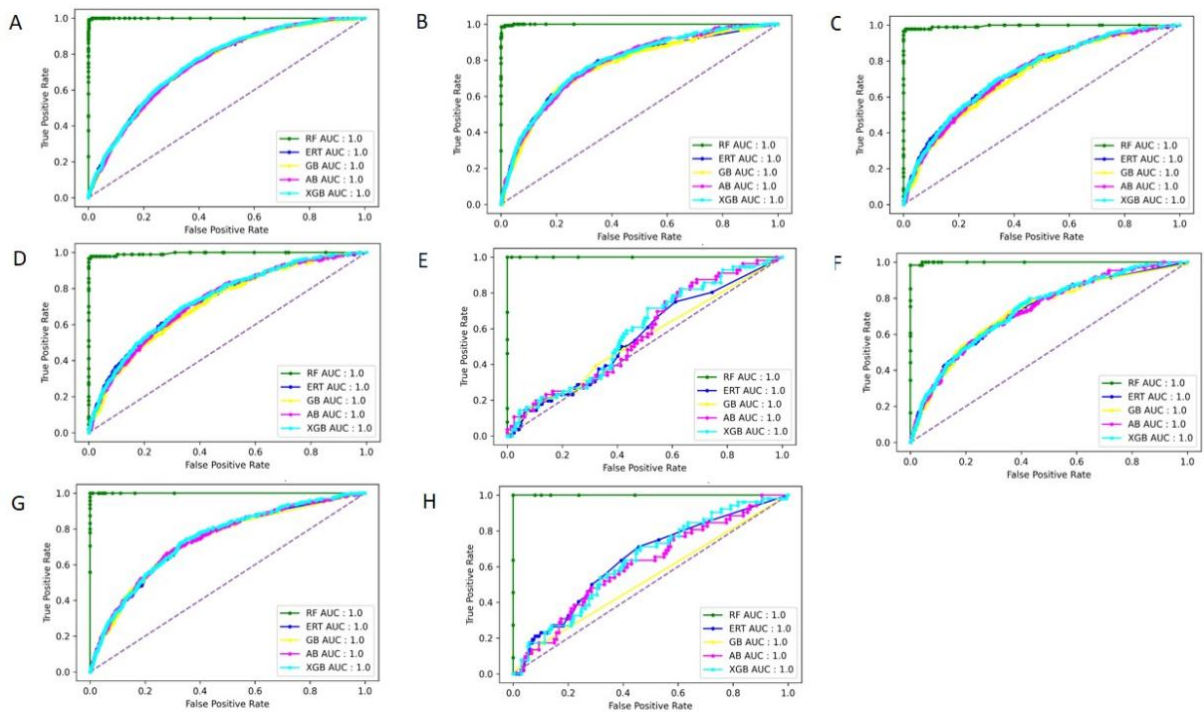

*Figure 10 ROC curve of ML Classifiers on 10-fold cross-validation independent test sets, A: E. coli, B: B. velezensis, C: Bacilus subtilis, D: C. glutamicum, E: G. kaustophilus, F: M. tuber, G: S. eriocheiris, H: S. typhimurium*

*Table 5: Performance comparison of the proposed feature on 5-fold cross-validation set using different classifiers including Long short term memory (LSTM) and ML classifiers Random Forest (RF), Extreme Random Tree (ERT), Gradient Boosting (GB), AdaBoost (AB), Extreme Gradient Boosting (XGB)*

| Species | Classifier | ACC | SN | SP | MCC | AUC | F1 |
|---|---|---|---|---|---|---|---|
| *Average of all species* | LSTM | 0.79 | 0.88 | 0.35 | 0.25 | 0.62 | 0.87 |
| | RF | 1 | 1 | 1 | 1 | 1 | 1 |
| | ERT | 0.97 | 0.99 | 0.84 | 0.87 | 1 | 0.98 |
| | GB | 0.96 | 0.98 | 0.85 | 0.83 | 1 | 0.97 |
| | AB | 0.97 | 0.99 | 0.88 | 0.87 | 1 | 0.98 |
| | XGB | 0.97 | 0.99 | 0.87 | 0.87 | 1 | 0.98 |

5-fold cross-validation graphs for every species with their respective ROC curve for each classifiers can be seen in Figure 11 below.

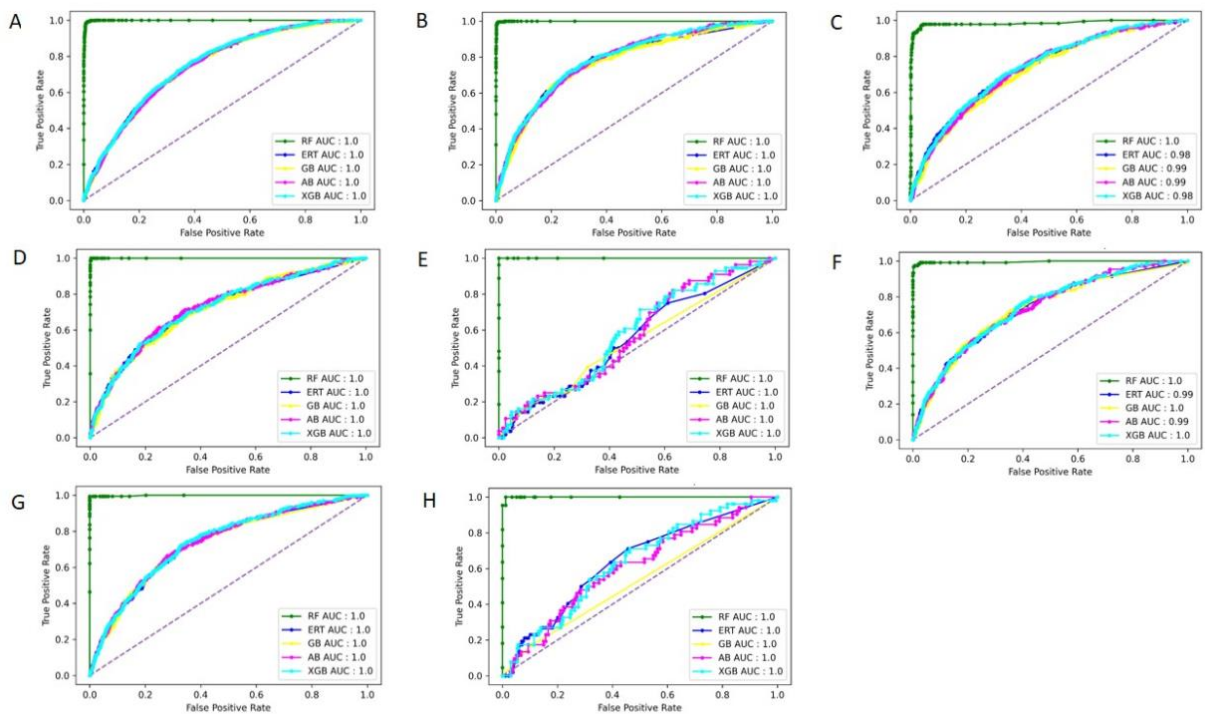

*Figure 11 ROC curve of ML Classifiers on 5-fold cross-validation set, A: E. coli, B: B. velezensis, C: Bacilus subtilis, D: C. glutamicum, E: G. kaustophilus, F: M. tuber, G: S. eriocheiris, H: S. typhimurium*

## 6.5. Comparison and Discussion

We compared our method for K-Ace sequence prediction with STALLION [37]. They used eleven different encoding methods which represented different characteristics of sequence and stacked them to further use three types of feature selection methods. They designed five tree-based ensemble models at the end which were built as specific for each species. Where our method uses deep learning model LSTM for feature extraction, we used the same classifiers used by STALLION to compare our results. The classification is done using ML classifiers RF, ERT, GB, AB, and XGB. As shown in Table 6 for training comparison for specie B. subtilis our method gives on average AUC of 0.958 while STALLION has 0.8948 similarly for other species like C. glutamicum, E. coli, G. kaustophilus, M. tuberculosis, and S. typhimurium our AUC are respectively 1.0, 0.994, 1, 0.998 and 0.984 while STALLION has 0.901, 0.8876, 0.888 and 0.895. In terms of Average ACC for species B. subtilis, C. glutamicum, E. coli, G. kaustophilus, M. tuberculosis, and S. typhimurium scores are 0.9360, 1, 0.9742, 0.9972, 0.9922 and 0.9933 and

STALLION has 0.7992, 0.8092, 0.7908, 0.8148, 0.8324 and 0.7834. Other metrics in Table 6 also show that our model was able to predict K-Ace sequences better than STALLION.

*Table 6: Performance comparison of the proposed feature on training set with previous model using ML classifiers Random Forest (RF), Extreme Random Tree (ERT), Gradient Boosting (GB), AdaBoost (AB), Extreme Gradient Boosting (XGB)*

| Species | Model | Classifier | ACC | Sn | Sp | MCC | AUC |
|---|---|---|---|---|---|---|---|
| *B. subtilis* | STALLION | RF | 0.799 | 0.757 | 0.84 | 0.599 | 0.907 |
| | | ERT | 0.802 | 0.833 | 0.771 | 0.605 | 0.903 |
| | | GB | 0.791 | 0.778 | 0.805 | 0.583 | 0.857 |
| | | AB | 0.805 | 0.701 | 0.908 | 0.623 | 0.908 |
| | | XGB | 0.799 | 0.597 | 1 | 0.652 | 0.899 |
| | **Deep-Ace** | **RF** | 1 | 1 | 1 | 1 | 1 |
| | | **ERT** | 1 | 1 | 1 | 1 | 1 |
| | | **GB** | 0.896 | 0.965 | 0.595 | 0.627 | 0.92 |
| | | **AB** | 0.886 | 0.955 | 0.583 | 0.592 | 0.93 |
| | | **XGB** | 0.9 | 0.97 | 0.591 | 0.64 | 0.94 |
| *C. glutamicum* | STALLION | RF | 0.808 | 0.816 | 0.8 | 0.616 | 0.915 |
| | | ERT | 0.799 | 0.786 | 0.813 | 0.599 | 0.917 |
| | | GB | 0.795 | 0.806 | 0.784 | 0.59 | 0.828 |
| | | AB | 0.823 | 0.844 | 0.801 | 0.646 | 0.922 |
| | | XGB | 0.821 | 0.886 | 0.757 | 0.648 | 0.923 |
| | **Deep-Ace** | **RF** | 1 | 1 | 1 | 1 | 1 |
| | | **ERT** | 1 | 1 | 1 | 1 | 1 |
| | | **GB** | 1 | 1 | 1 | 1 | 1 |
| | | **AB** | 1 | 1 | 1 | 1 | 1 |
| | | **XGB** | 1 | 1 | 1 | 1 | 1 |
| *E. coli* | STALLION | RF | 0.791 | 0.776 | 0.806 | 0.582 | 0.905 |
| | | ERT | 0.789 | 0.806 | 0.772 | 0.578 | 0.905 |
| | | GB | 0.788 | 0.78 | 0.797 | 0.576 | 0.824 |
| | | AB | 0.804 | 0.755 | 0.852 | 0.61 | 0.909 |
| | | XGB | 0.782 | 0.565 | 1 | 0.627 | 0.895 |
| | **Deep-Ace** | **RF** | 1 | 1 | 1 | 1 | 1 |
| | | **ERT** | 1 | 1 | 1 | 1 | 1 |
| | | **GB** | 0.958 | 0.97 | 0.934 | 0.905 | 0.99 |
| | | **AB** | 0.953 | 0.966 | 0.929 | 0.895 | 0.99 |
| | | **XGB** | 0.961 | 0.973 | 0.938 | 0.912 | 0.99 |
| *G. kaustophilus* | STALLION | RF | 0.806 | 0.796 | 0.816 | 0.612 | 0.908 |
| | | ERT | 0.816 | 0.811 | 0.82 | 0.631 | 0.912 |
| | | GB | 0.813 | 0.801 | 0.825 | 0.626 | 0.835 |
| | | AB | 0.823 | 0.816 | 0.83 | 0.646 | 0.866 |
| | | XGB | 0.816 | 0.82 | 0.811 | 0.631 | 0.919 |
| | **Deep-Ace** | **RF** | 1 | 1 | 1 | 1 | 1 |
| | | **ERT** | 1 | 1 | 1 | 1 | 1 |
| | | **GB** | 0.987 | 0.993 | 0.956 | 0.951 | 1 |
| | | **AB** | 1 | 1 | 1 | 1 | 1 |
| | | **XGB** | 1 | 1 | 1 | 1 | 1 |
| *M. tuberculosis* | STALLION | RF | 0.831 | 0.82 | 0.843 | 0.663 | 0.928 |
| | | ERT | 0.826 | 0.836 | 0.816 | 0.652 | 0.917 |
| | | GB | 0.818 | 0.809 | 0.827 | 0.636 | 0.853 |
| | | AB | 0.842 | 0.806 | 0.878 | 0.685 | 0.932 |
| | | XGB | 0.845 | 0.791 | 0.899 | 0.694 | 0.845 |
| | **Deep-Ace** | **RF** | 1 | 1 | 1 | 1 | 1 |
| | | **ERT** | 1 | 1 | 1 | 1 | 1 |
| | | **GB** | 0.975 | 0.988 | 0.919 | 0.917 | 0.99 |
| | | **AB** | 0.995 | 0.999 | 0.981 | 0.984 | 1 |
| | | **XGB** | 0.992 | 0.998 | 0.967 | 0.973 | 1 |
| *S. typhimurium* | STALLION | RF | 0.77 | 0.768 | 0.773 | 0.54 | 0.893 |
| | | ERT | 0.775 | 0.783 | 0.768 | 0.551 | 0.843 |
| | | GB | 0.775 | 0.768 | 0.783 | 0.551 | 0.813 |
| | | AB | 0.796 | 0.788 | 0.803 | 0.591 | 0.906 |
| | | XGB | 0.801 | 0.823 | 0.778 | 0.602 | 0.907 |
|  | Deep-Ace | RF | 1 | 1 | 1 | 1 | 1 |
|  |  | ERT | 1 | 1 | 1 | 1 | 1 |
|  |  | GB | 0.967 | 0.98 | 0.863 | 0.833 | 0.92 |
|  |  | AB | 1 | 1 | 1 | 1 | 1 |
|  |  | XGB | 1 | 1 | 1 | 1 | 1 |
| Average of all Classifiers | STALLION | RF | 0.801 | 0.789 | 0.813 | 0.602 | 0.91 |
|  |  | ERT | 0.802 | 0.81 | 0.794 | 0.603 | 0.9 |
|  |  | GB | 0.797 | 0.791 | 0.804 | 0.594 | 0.835 |
|  |  | AB | 0.816 | 0.785 | 0.846 | 0.634 | 0.908 |
|  |  | XGB | 0.811 | 0.747 | 0.875 | 0.643 | 0.898 |
|  | Deep-Ace | RF | 1 | 1 | 1 | 1 | 1 |
|  |  | ERT | 1 | 1 | 1 | 1 | 1 |
|  |  | GB | 0.97 | 0.985 | 0.9 | 0.894 | 0.977 |
|  |  | AB | 0.978 | 0.989 | 0.932 | 0.929 | 0.99 |
|  |  | XGB | 0.981 | 0.992 | 0.933 | 0.936 | 0.992 |

To perform a true comparison we have evaluated our results on independent testing as well shown in Table 7 independent test comparison for specie B. subtilis our method gives on average AUC 0.73 while STALLION has 0.6016 similarly for other species like C. glutamicum, E. coli, G. kaustophilus, M. tuberculosis and S. typhimurium our AUC are respectively 0.72, 0.748, 0.564, 0.732 and 0.616 while STALLION has 0.665, 0.6208, 0.597, 0.6704 and 0.6194. In terms of Average ACC for species B. subtilis, C. glutamicum, E. coli, G. kaustophilus, M. tuberculosis, and S. typhimurium are scores are 0.7955, 0.7832, 0.70558, 0.7427, 0.7902 and 0.8286 and STALLION has 0.6102, 0.599, 0.5916, 0.4796, 0.6412 and 0.534. Other metrics in Table 7 also show that our model during independent testing out performers STALLION using the same ML classifiers.

*Table 7: Performance comparison of the proposed feature on Independent set with previous model using ML classifiers Random Forest (RF), Extreme Random Tree (ERT), Gradient Boosting (GB), AdaBoost (AB), Extreme Gradient Boosting (XGB)*

| Species | Model | Classifier | ACC | Sn | Sp | MCC | AUC |
|---|---|---|---|---|---|---|---|
|  | STALLION | RF | 0.573 | 0.608 | 0.569 | 0.105 | 0.63 |

| Species | Method | Model | | | | | |
|---|---|---|---|---|---|---|---|
| *B. subtilis* | | ERT | 0.568 | 0.616 | 0.563 | 0.106 | 0.612 |
| | | GB | 0.558 | 0.632 | 0.55 | 0.108 | 0.562 |
| | | AB | 0.65 | 0.504 | 0.666 | 0.106 | 0.623 |
| | | XGB | 0.702 | 0.368 | 0.737 | 0.07 | 0.581 |
| | **Deep-Ace** | **RF** | 0.804 | 0.927 | 0.292 | 0.27 | 0.73 |
| | | **ERT** | 0.809 | 0.939 | 0.27 | 0.271 | 0.73 |
| | | **GB** | 0.785 | 0.894 | 0.331 | 0.249 | 0.72 |
| | | **AB** | 0.787 | 0.89 | 0.356 | 0.267 | 0.73 |
| | | **XGB** | 0.796 | 0.909 | 0.326 | 0.27 | 0.74 |
| *C. glutamicum* | STALLION | RF | 0.61 | 0.687 | 0.602 | 0.168 | 0.69 |
| | | ERT | 0.61 | 0.687 | 0.602 | 0.168 | 0.676 |
| | | GB | 0.59 | 0.627 | 0.587 | 0.124 | 0.608 |
| | | AB | 0.607 | 0.759 | 0.592 | 0.203 | 0.678 |
| | | XGB | 0.578 | 0.783 | 0.558 | 0.196 | 0.673 |
| | **Deep-Ace** | **RF** | 0.781 | 0.88 | 0.378 | 0.271 | 0.71 |
| | | **ERT** | 0.787 | 0.886 | 0.382 | 0.285 | 0.72 |
| | | **GB** | 0.781 | 0.875 | 0.399 | 0.284 | 0.72 |
| | | **AB** | 0.787 | 0.882 | 0.399 | 0.295 | 0.73 |
| | | **XGB** | 0.782 | 0.877 | 0.396 | 0.284 | 0.72 |
| *E. coli* | STALLION | RF | 0.559 | 0.654 | 0.534 | 0.152 | 0.631 |
| | | ERT | 0.559 | 0.651 | 0.535 | 0.151 | 0.624 |
| | | GB | 0.543 | 0.645 | 0.517 | 0.131 | 0.582 |
| | | AB | 0.622 | 0.693 | 0.604 | 0.241 | 0.657 |
| | | XGB | 0.675 | 0.424 | 0.741 | 0.146 | 0.61 |
| | **Deep-Ace** | **RF** | 0.708 | 0.795 | 0.536 | 0.336 | 0.75 |
| | | **ERT** | 0.709 | 0.799 | 0.531 | 0.336 | 0.75 |
| | | **GB** | 0.704 | 0.783 | 0.549 | 0.333 | 0.74 |
| | | **AB** | 0.7 | 0.778 | 0.547 | 0.325 | 0.75 |
| | | **XGB** | 0.708 | 0.786 | 0.556 | 0.343 | 0.75 |
| *G. kaustophilus* | STALLION | RF | 0.502 | 0.706 | 0.484 | 0.104 | 0.63 |
| | | ERT | 0.479 | 0.647 | 0.464 | 0.061 | 0.625 |
| | | GB | 0.474 | 0.647 | 0.458 | 0.058 | 0.582 |
| | | AB | 0.445 | 0.588 | 0.432 | 0.011 | 0.555 |
| | | XGB | 0.498 | 0.647 | 0.484 | 0.072 | 0.593 |
| | **Deep-Ace** | **RF** | 0.749 | 0.854 | 0.215 | 0.069 | 0.58 |
| | | **ERT** | 0.749 | 0.857 | 0.197 | 0.055 | 0.55 |
| | | **GB** | 0.737 | 0.84 | 0.215 | 0.053 | 0.54 |
| | | **AB** | 0.74 | 0.836 | 0.25 | 0.083 | 0.57 |
| | | **XGB** | 0.74 | 0.843 | 0.215 | 0.057 | 0.58 |
| *M. tuberculosis* | STALLION | RF | 0.646 | 0.735 | 0.634 | 0.241 | 0.681 |
| | | ERT | 0.641 | 0.735 | 0.628 | 0.237 | 0.666 |
| | | GB | 0.612 | 0.721 | 0.597 | 0.206 | 0.667 |
| | | AB | 0.656 | 0.706 | 0.65 | 0.234 | 0.689 |
| | | XGB | 0.651 | 0.647 | 0.652 | 0.197 | 0.649 |
| | **Deep-Ace** | **RF** | 0.794 | 0.892 | 0.371 | 0.282 | 0.73 |
| | | **ERT** | 0.796 | 0.9 | 0.348 | 0.274 | 0.73 |
| | | **GB** | 0.784 | 0.877 | 0.383 | 0.268 | 0.73 |
| | | **AB** | 0.791 | 0.881 | 0.402 | 0.292 | 0.73 |
| | | **XGB** | 0.788 | 0.88 | 0.394 | 0.283 | 0.74 |
| *S. typhimurium* | STALLION | RF | 0.542 | 0.7 | 0.535 | 0.096 | 0.615 |
| | | ERT | 0.564 | 0.7 | 0.558 | 0.106 | 0.645 |
| | | GB | 0.537 | 0.7 | 0.53 | 0.095 | 0.615 |
| | | AB | 0.52 | 0.6 | 0.516 | 0.048 | 0.619 |
| | | XGB | 0.507 | 0.5 | 0.507 | 0.003 | 0.603 |
| | **Deep-Ace** | **RF** | 0.848 | 0.944 | 0.154 | 0.127 | 0.67 |
| | | **ERT** | 0.843 | 0.942 | 0.135 | 0.099 | 0.64 |
| | | **GB** | 0.803 | 0.891 | 0.174 | 0.065 | 0.53 |
| | | **AB** | 0.815 | 0.907 | 0.154 | 0.066 | 0.61 |
| | | **XGB** | 0.836 | 0.928 | 0.174 | 0.119 | 0.63 |
| *Average of all classifiers* | STALLION | **RF** | 0.572 | 0.682 | 0.56 | 0.145 | 0.647 |
| | | **ERT** | 0.571 | 0.673 | 0.559 | 0.139 | 0.642 |
| | | **GB** | 0.553 | 0.662 | 0.54 | 0.121 | 0.603 |

| | | AB | 0.584 | 0.642 | 0.577 | 0.141 | 0.637 |
| | | XGB | 0.602 | 0.562 | 0.614 | 0.114 | 0.619 |
| | Deep-Ace | RF | 0.783 | 0.881 | 0.354 | 0.252 | 0.708 |
| | | ERT | 0.783 | 0.884 | 0.341 | 0.244 | 0.705 |
| | | GB | 0.77 | 0.861 | 0.373 | 0.239 | 0.685 |
| | | AB | 0.772 | 0.862 | 0.377 | 0.245 | 0.705 |
| | | XGB | 0.777 | 0.869 | 0.37 | 0.249 | 0.712 |

Table 8 shows a species-based comparison with STALLION and for each specie Deep-Ace have performed better and given state of art results for *B. subtilis, C. glutamicum, E. coli, G. kaustophilus, M. tuberculosis*, and *S. typhimurium* with ACC 0.7955, 0.7832, 0.7055, 0.7427, 0.7902, and 0.8286. Similarly, when evaluated on other measures such as AUC our model has outperformed STALLION with AUC 0.73, 0.72, 0.748, 0.564, 0.732 and 0.616. This makes Deep-Ace is a reliable method for specie-based prediction of K-Ace and non-K-Ace sites.

*Table 8: Independent testing comparison of the average of all species with previous model*

| Species | Model | ACC | Sn | Sp | MCC | AUC |
|---|---|---|---|---|---|---|
| *B. subtilis* | **STALLION** | 0.62 | 0.55 | 0.62 | 0.1 | 0.61 |
| | **Deep-Ace** | 0.8 | 0.92 | 0.32 | 0.27 | 0.73 |
| *C. glutamicum* | **STALLION** | 0.6 | 0.71 | 0.59 | 0.18 | 0.67 |
| | **Deep-Ace** | 0.79 | 0.88 | 0.4 | 0.29 | 0.72 |
| *E. coli* | **STALLION** | 0.6 | 0.62 | 0.59 | 0.17 | 0.63 |
| | **Deep-Ace** | 0.71 | 0.79 | 0.55 | 0.34 | 0.75 |
| *G. kaustophilus* | **STALLION** | 0.48 | 0.65 | 0.47 | 0.07 | 0.6 |
| | **Deep-Ace** | 0.75 | 0.85 | 0.22 | 0.07 | 0.57 |
| *M. tuberculosis* | **STALLION** | 0.65 | 0.71 | 0.64 | 0.23 | 0.68 |
| | **Deep-Ace** | 0.8 | 0.89 | 0.38 | 0.28 | 0.74 |
| *S. typhimurium* | **STALLION** | 0.54 | 0.64 | 0.53 | 0.07 | 0.62 |
| | **Deep-Ace** | 0.83 | 0.93 | 0.16 | 0.1 | 0.62 |

## 6.6. Feature Space Visualization Comparison

In this section, a discussion on deep feature representations in 2-D space is presented which are generated using the T-Stochastic Neighbor Embedding (T-SNE) algorithm[84]. T-SNE is a statistical method to graphically visualize high-dimensional data and map the crowd data points in two or three-dimensional maps by reducing the chances of crowded points being together in the center. The feature space visualizations of deep representations are shown in Figure 12 it can be easily seen that the feature space boundary is clearly defined samples of data (i.e. between positive and negative samples) and the model can separately identify the classes with relatively less effort.

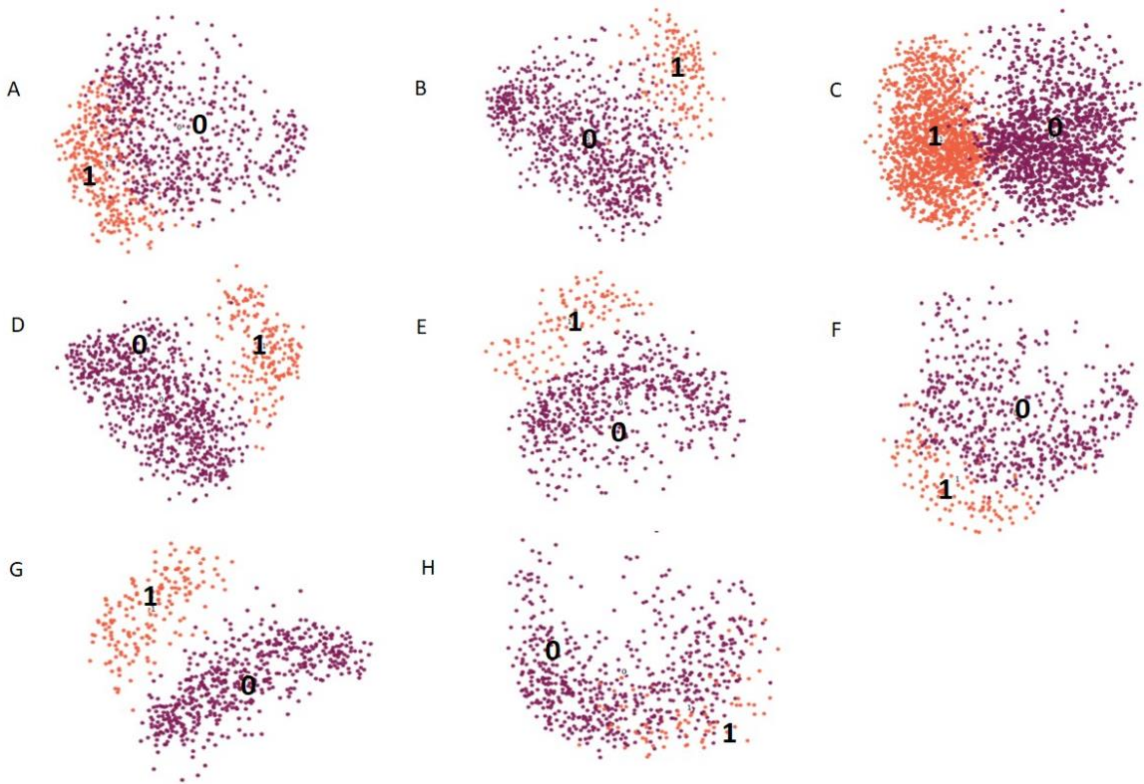

*Figure 12 TSNE Visualization of LSTM representations of different species, A: E. coli, B: B. velezensis, C: Bacilus subtilis, D: C. glutamicum, E: G. kaustophilus, F: M. tuber, G: S. eriocheiris, H: S. typhimurium*

## 7. Conclusion

This study proposed Deep-Ace as a deep learning-based framework for K-Ace site prediction for eight different prokaryotic species. The dysregulation of K-Ace has been linked to several diseases like cancer, cardiovascular and immune disorders, aging factors, and neurological diseases. It carries importance in cell biology making it important to have a better understanding of its modularity. Deep-Ace is based on LSTM which was used to extract features using raw input of K-

Ace and non K-Ace sequences. The features extracted through the network when given to different classifiers outperform previous methods which used hand-crafted and encoding-based features. As deep learning-based features capture every aspect of our sequences as seen by feature representations it is a more efficient method. Hence K-Ace site prediction for eight different species in our model has shown state-of-the-art results moreover our study identifies heterogeneous and complementary features derived from different perspectives that helped to improve predictor performance.

# References

1. Ramazi, S. and J. Zahiri, *Post-translational modifications in proteins: resources, tools and prediction methods.* Database 2021. baab012.
2. Soppa, J.J.A., *Protein acetylation in archaea, bacteria, and eukaryotes.* 2010.
3. Kouzarides, T., *Acetylation: a regulatory modification to rival phosphorylation?.* The EMBO journal, 2000. **19**(6): p. 1176-1179.
4. Zhao, S., et al., *Regulation of cellular metabolism by protein lysine acetylation.* Science 2010. **327**(5968): p. 1000-1004.
5. Xia, C., et al., *Protein acetylation and deacetylation: An important regulatory modification in gene transcription.* 2020. **20**(4): p. 2923-2940.
6. Tang, Y., et al., *Acetylation is indispensable for p53 activation.* 2008. **133**(4): p. 612-626.
7. Zhang, J., et al., *Lysine acetylation is a highly abundant and evolutionarily conserved modification in Escherichia coli.* 2009. **8**(2): p. 215-225.
8. Brooks, C.L., W.J.P. Gu, and cell, *The impact of acetylation and deacetylation on the p53 pathway.* 2011. **2**(6): p. 456-462.
9. Hubbert, C., et al., *HDAC6 is a microtubule-associated deacetylase.* 2002. **417**(6887): p. 455-458.
10. Schlesinger, J., et al., *The cardiac transcription network modulated by Gata4, Mef2a, Nkx2. 5, Srf, histone modifications, and microRNAs.* 2011. **7**(2): p. e1001313.
11. Gallinari, P., et al., *HDACs, histone deacetylation and gene transcription: from molecular biology to cancer therapeutics.* 2007. **17**(3): p. 195-211.
12. Kim, D., et al., *SIRT1 deacetylase protects against neurodegeneration in models for Alzheimer's disease and amyotrophic lateral sclerosis.* 2007. **26**(13): p. 3169-3179.
13. Fraga, M.F., et al., *Loss of acetylation at Lys16 and trimethylation at Lys20 of histone H4 is a common hallmark of human cancer.* 2005. **37**(4): p. 391-400.
14. Medzihradszky, K.F.J.M.i.e., *Peptide sequence analysis.* 2005. **402**: p. 209-244.
15. Deng, W., et al., *GPS-PAIL: prediction of lysine acetyltransferase-specific modification sites from protein sequences.* 2016. **6**(1): p. 1-10.
16. Yu, K., et al., *Deep learning based prediction of reversible HAT/HDAC-specific lysine acetylation.* 2020. **21**(5): p. 1798-1805.
17. Yang, Y., et al., *Prediction and analysis of multiple protein lysine modified sites based on conditional wasserstein generative adversarial networks.* 2021. **22**(1): p. 1-17.
18. Xiu, Q., et al. *Prediction method for lysine acetylation sites based on LSTM network.* in *IEEE International Conference on Computer Science and Network Technology (ICCSNT).* 2019. IEEE.
19. Basith, S., et al., *Recent trends on the development of machine learning approaches for the prediction of lysine acetylation sites.* 2022. **29**(2): p. 235-250.
20. Li, A., et al., *Prediction of Nε-acetylation on internal lysines implemented in Bayesian Discriminant Method.* 2006. **350**(4): p. 818-824.
21. Li, S., et al., *Improved prediction of lysine acetylation by support vector machines.* 2009. **16**(8): p. 977-983.
22. Xu, Y., et al., *Lysine acetylation sites prediction using an ensemble of support vector machine classifiers.* 2010. **264**(1): p. 130-135.
23. Lee, T.Y., et al., *N-Ace: Using solvent accessibility and physicochemical properties to identify protein N-acetylation sites.* 2010. **31**(15): p. 2759-2771.

24. Shao, J., et al., *Systematic analysis of human lysine acetylation proteins and accurate prediction of human lysine acetylation through bi-relative adapted binomial score Bayes feature representation.* 2012. **8**(11): p. 2964-2973.
25. Shi, S.-P., et al., *PLMLA: prediction of lysine methylation and lysine acetylation by combining multiple features.* 2012. **8**(5): p. 1520-1527.
26. Suo, S.-B., et al., *Position-specific analysis and prediction for protein lysine acetylation based on multiple features.* 2012. **7**(11): p. e49108.
27. Suo, S.-B., et al., *Proteome-wide analysis of amino acid variations that influence protein lysine acetylation.* 2013. **12**(2): p. 949-958.
28. Hou, T., et al., *LAceP: lysine acetylation site prediction using logistic regression classifiers.* 2014. **9**(2): p. e89575.
29. Lu, C.-T., et al., *An intelligent system for identifying acetylated lysine on histones and nonhistone proteins.* 2014. **2014**.
30. Li, Y., et al., *Accurate in silico identification of species-specific acetylation sites by integrating protein sequence-derived and functional features.* 2014. **4**(1): p. 1-12.
31. Qiu, W.-R., et al., *iPTM-mLys: identifying multiple lysine PTM sites and their different types.* 2016. **32**(20): p. 3116-3123.
32. Wuyun, Q., et al., *Improved species-specific lysine acetylation site prediction based on a large variety of features set.* 2016. **11**(5): p. e0155370.
33. Chen, G., et al., *ProAcePred: prokaryote lysine acetylation sites prediction based on elastic net feature optimization.* 2018. **34**(23): p. 3999-4006.
34. Chen, G., et al., *Prediction and functional analysis of prokaryote lysine acetylation site by incorporating six types of features into Chou's general PseAAC.* 2019. **461**: p. 92-101.
35. Ning, Q., et al., *Analysis and prediction of human acetylation using a cascade classifier based on support vector machine.* 2019. **20**(1): p. 1-15.
36. Yu, B., et al., *DNNAce: prediction of prokaryote lysine acetylation sites through deep neural networks with multi-information fusion.* 2020. **200**: p. 103999.
37. Basith, S., G. Lee, and B.J.B.i.b. Manavalan, *STALLION: a stacking-based ensemble learning framework for prokaryotic lysine acetylation site prediction.* 2022. **23**(1): p. bbab376.
38. Shahid, M., et al., *ORI-Deep: improving the accuracy for predicting origin of replication sites by using a blend of features and long short-term memory network.* Briefings in Bioinformatics, 2022.
39. Malebary, S.J. and Y.D. Khan, *Evaluating machine learning methodologies for identification of cancer driver genes.* Scientific reports, 2021. **11**(1): p. 1-13.
40. Malebary, S.J., E. Alzahrani, and Y.D. Khan, *A comprehensive tool for accurate identification of methyl-glutamine sites.* Journal of Molecular Graphics and Modelling, 2022. **110**: p. 108074.
41. Hussain, W., *sAMP-PFPDeep: Improving accuracy of short antimicrobial peptides prediction using three different sequence encodings and deep neural networks.* Briefings in Bioinformatics, 2022. **23**(1): p. bbab487.
42. Baig, T.I., et al., *Ilipo-pseaac: identification of lipoylation sites using statistical moments and general pseaac.* Computers, Materials and Continua, 2022. **71**(1): p. 215-230.
43. Alghamdi, W., et al., *LBCEPred: a machine learning model to predict linear B-cell epitopes.* Briefings in Bioinformatics, 2022. 23(3): bbac035.

44. Xu, H., et al., *PLMD: an updated data resource of protein lysine modifications.* 2017. **44**(5): p. 243-250.
45. Li, W. and A.J.B. Godzik, *Cd-hit: a fast program for clustering and comparing large sets of protein or nucleotide sequences.* 2006. **22**(13): p. 1658-1659.
46. Shahid, M., et al., *ORI-Deep: improving the accuracy for predicting origin of replication sites by using a blend of features and long short-term memory network.* 2022. **23**(2): p. bbac001.
47. Roy, D., K.S.R. Murty, and C.K. Mohan. *Feature selection using deep neural networks.* in *2015 International Joint Conference on Neural Networks (IJCNN).* 2015. IEEE.
48. Yousoff, S.N.M., A. Baharin, and A. Abdullah. *A review on optimization algorithm for deep learning method in bioinformatics field.* in *2016 IEEE EMBS Conference on Biomedical Engineering and Sciences (IECBES).* 2016. IEEE.
49. Lim, A., S. Lim, and S. Kim, *Enhancer prediction with histone modification marks using a hybrid neural network model.* Methods, 2019. **166**: p. 48-56.
50. Naseer, S., et al., *Optimization of serine phosphorylation prediction in proteins by comparing human engineered features and deep representations.* Analytical Biochemistry, 2021. **615**: p. 114069.
51. Sherstinsky, A.J.P.D.N.P., *Fundamentals of recurrent neural network (RNN) and long short-term memory (LSTM) network.* 2020. **404**: p. 132306.
52. Hochreiter, S. and J.J.A.i.n.i.p.s. Schmidhuber, *LSTM can solve hard long time lag problems.* 1996. **9**.
53. Hochreiter, S. and J. Schmidhuber, *Long short-term memory.* Neural computation, 1997. **9**(8): p. 1735-1780.
54. Breuel, T.M.J.a.p.a., *Benchmarking of LSTM networks.* 2015.
55. Gers, F.A. and J. Schmidhuber. *Recurrent nets that time and count.* in *Proceedings of the IEEE-INNS-ENNS International Joint Conference on Neural Networks. IJCNN 2000.*
56. Bishop, C.M. and N.M. Nasrabadi, *Pattern recognition and machine learning.* Vol. 4. 2006: Springer.
57. Ho, T.K., *The random subspace method for constructing decision forests.* IEEE transactions on pattern analysis and machine intelligence, 1998. **20**(8): p. 832-844.
58. Malebary, S., et al., *iAcety–SmRF: Identification of Acetylation Protein by Using Statistical Moments and Random Forest.* 2022. **12**(3): p. 265.
59. You, X., et al., *Recognizing protein-metal ion ligands binding residues by random forest algorithm with adding orthogonal properties.* 2022. **98**: p. 107693.
60. Chen, C., et al., *Improving protein-protein interactions prediction accuracy using XGBoost feature selection and stacked ensemble classifier.* 2020. **123**: p. 103899.
61. Geurts, P., D. Ernst, and L.J.M.l. Wehenkel, *Extremely randomized trees.* 2006. **63**(1): p. 3-42.
62. Lin, J., Y.J.P. Wang, and P. Letters, *Using a novel AdaBoost algorithm and Chou's pseudo amino acid composition for predicting protein subcellular localization.* 2011. **18**(12): p. 1219-1225.
63. Niu, B., et al., *Using AdaBoost for the prediction of subcellular location of prokaryotic and eukaryotic proteins.* 2008. **12**(1): p. 41-45.
64. Friedman, J.H.J.A.o.s., *Greedy function approximation: a gradient boosting machine.* 2001: p. 1189-1232.

65. Wang, H., C. Liu, and L.J.S.r. Deng, *Enhanced prediction of hot spots at protein-protein interfaces using extreme gradient boosting.* 2018. **8**(1): p. 1-13.
66. Zhao, Z., et al., *Identify DNA-binding proteins through the extreme gradient boosting algorithm.* 2022. **12**: p. 821996.
67. Yu, B., et al., *SubMito-XGBoost: predicting protein submitochondrial localization by fusing multiple feature information and eXtreme gradient boosting.* 2020. **36**(4): p. 1074-1081.
68. Sikander, R., A. Ghulam, and F.J.S.R. Ali, *XGB-DrugPred: computational prediction of druggable proteins using eXtreme gradient boosting and optimized features set.* 2022. **12**(1): p. 1-9.
69. Arif, M., et al., *StackACPred: Prediction of anticancer peptides by integrating optimized multiple feature descriptors with stacked ensemble approach.* Chemometrics and Intelligent Laboratory Systems, 2021: p. 104458.
70. Alzahrani, E., et al., *Identification of stress response proteins through fusion of machine learning models and statistical paradigms.* Scientific Reports, 2021. **11**(1): p. 1-15.
71. Hamel, L., *Model assessment with ROC curves*, in *Encyclopedia of Data Warehousing and Mining, Second Edition*. 2009, IGI Global. p. 1316-1323.
72. Lipton, Z.C., C. Elkan, and B.J.a.p.a. Narayanaswamy, *Thresholding classifiers to maximize F1 score.* 2014.
73. Zhu, Q.J.P.R.L., *On the performance of Matthews correlation coefficient (MCC) for imbalanced dataset.* 2020. **136**: p. 71-80.
74. Refaeilzadeh, P., L. Tang, and H.J.E.o.d.s. Liu, *Cross-validation.* 2009. **5**: p. 532-538.
75. Barukab, O., et al., *iSulfoTyr-PseAAC: Identify tyrosine sulfation sites by incorporating statistical moments via Chou's 5-steps rule and pseudo components.* Current Genomics, 2019. **20**(4): p. 306-320.
76. Malebary, S.J., R. Khan, and Y.D. Khan, *ProtoPred: advancing oncological research through identification of proto-oncogene proteins.* IEEE Access, 2021. **9**: p. 68788-68797.
77. Ilyas, S., et al., *iMethylK-PseAAC: Improving accuracy of lysine methylation sites identification by incorporating statistical moments and position relative features into general PseAAC via Chou's 5-steps rule.* Current Genomics, 2019. **20**(4): p. 275-292.
78. Allehaibi, K., Y. Daanial Khan, and S.A. Khan, *iTAGPred: A Two-Level Prediction Model for Identification of Angiogenesis and Tumor Angiogenesis Biomarkers.* Applied Bionics and Biomechanics, 2021. **2021**.
79. Van der Maaten, L. and G. Hinton, *Visualizing data using t-SNE.* Journal of machine learning research, 2008. **9**(11).